\documentclass{article}

\usepackage{microtype}
\usepackage{graphicx}
\usepackage{subfigure}
\usepackage{booktabs}

\usepackage{hyperref}

\usepackage[accepted]{icml2019}

\usepackage{enumitem}

\graphicspath{ {./images/} }

\icmltitlerunning{MNIST-C}

\begin{document}

\twocolumn[
\icmltitle{MNIST-C: A Robustness Benchmark for Computer Vision}

\icmlsetsymbol{equal}{*}

\begin{icmlauthorlist}
\icmlauthor{Norman Mu}{goo}
\icmlauthor{Justin Gilmer}{goo}

\end{icmlauthorlist}

\icmlaffiliation{goo}{Google Inc.}

\icmlcorrespondingauthor{Norman Mu}{normanmu@google.com}
\icmlcorrespondingauthor{Justin Gilmer}{gilmer@google.com}

\icmlkeywords{Machine Learning, ICML}

\vskip 0.3in
]

\printAffiliationsAndNotice{}  

\begin{abstract}
We introduce the MNIST-C dataset, a comprehensive suite of 15 corruptions applied to the MNIST test set, for benchmarking out-of-distribution robustness in computer vision. Through several experiments and visualizations we demonstrate that our corruptions significantly degrade performance of state-of-the-art computer vision models while preserving the semantic content of the test images. In contrast to the popular notion of adversarial robustness, our model-agnostic corruptions do not seek worst-case performance but are instead designed to be broad and diverse, capturing multiple failure modes of modern models. In fact, we find that several previously published adversarial defenses \emph{significantly degrade} robustness as measured by MNIST-C. We hope that our benchmark serves as a useful tool for future work in designing systems that are able to learn robust feature representations that capture the underlying semantics of the input.
\end{abstract}

\section{Introduction}
\label{submission}
Despite reports of superhuman performance on test datasets drawn from the same distribution as the training data, computer vision models still lag behind humans when evaluated on out-of-distribution (OOD) data \cite{dodge2017study}. For example, models lack robustness to small translations of the input \cite{azulay2018deep}, small adversarial perturbations \cite{szegedy2013intriguing, goodfellow2014explaining}, as well as commonly occurring image corruptions such as brightness, fog and various forms of blurring \cite{pei2017towards, hendrycks2018benchmarking}. Achieving robustness to distributional shift is an essential step for deploying models in complex, real-world settings where test data does not perfectly match the training distribution.

\begin{figure*}[h]
\centering
\label{fig:corruptions}
\includegraphics[width=\textwidth]{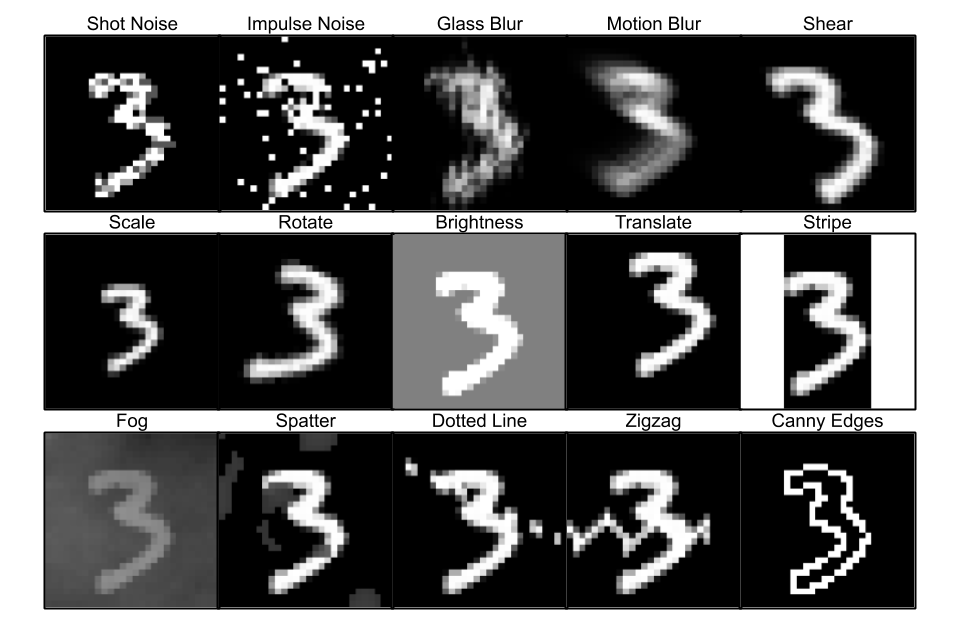}
\caption{Randomly sampled applications of all 15 corruptions comprising MNIST-C}
\end{figure*}

Recently, \cite{hendrycks2018benchmarking} proposed an OOD benchmark for the CIFAR-10 and Imagenet datasets. This benchmark consists of 15 commonly occurring visual corruptions at 5 different severity levels and is intended as a general-purpose robustness benchmark in computer vision. Smaller, simpler datasets such as MNIST continue to play an important role during prototyping, when iteration speed is paramount. MNIST is still commonly used in robustness research today \cite{madry2017towards,wang2018direct,frosst2018darccc,schott2018robust}; however, MNIST lacks a standardized corrupted variant. To this end we propose MNIST-C\footnote{The source code for MNIST-C and download link are available at \href{http://github.com/google-research/mnist-c}{http://github.com/google-research/mnist-c}.}, a benchmark consisting of 15 image corruptions for measuring out-of-distribution robustness in computer vision. Our benchmark is inspired by Imagenet-C and CIFAR-10-C, but is also specifically tailored to MNIST which consists of low resolution, black-and-white images. These 15 corruptions are carefully chosen out of a larger pool of 31 corruptions\footnote{While we recommend and publish results based on the subset of 15 for evaluation, we will open source code for generating all 31 for researchers to experiment with.}.

Through several experiments and visualizations, we demonstrate that our benchmark is non-trivial, semantically invariant, realistic, and diverse. Our corruptions capture failure modes of models previously unidentified in literature. Relative to clean data, our corruptions increase error rates of convolutional neural networks by a factor of 10 while preserving the semantic content of the underlying image. Furthermore, we evaluate 4 prior adversarial defense methods and find that they all significantly \emph{degrade} performance on MNIST-C. We believe this shows that our robustness benchmark captures failure modes of computer vision that popular measures of adversarial robustness fail to identify. Finally, we demonstrate that simple data augmentation cannot trivially solve this benchmark--- training on all but one of the corruptions yields minimal improvements on test accuracy for the held out corruption, leaving a large gap between neural network and human performance.

\section{MNIST-C}

Here, we briefly explain the 15 corruptions in MNIST-C.
\textbf{Shot noise}, \textbf{impulse noise} are random corruptions which may occur during the imaging process.
\textbf{Glass blur} simulates viewing the image through frosted glass by locally shuffling pixels.
\textbf{Motion blur} blurs the image along a random line.
\textbf{Shear}, \textbf{translate}, \textbf{scale}, \textbf{rotate} \cite{ghifary2015domain, engstrom2017rotation} are each applied as an affine transformation to the image.
\textbf{Brightness} increases the brightness of the image. 
\textbf{Stripe} inverts the pixel values along a vertical stripe in the center of the image.
\textbf{Fog} simulates a haze or fog using the diamond square algorithm.
\textbf{Spatter} occludes small regions of the image with randomly generated splotches.
\textbf{Zigzag}, \textbf{dotted line} superimpose randomly oriented zigzags and dotted lines over the image, with the brightness of each straight segment controlled by an exponential kernel.
\textbf{Canny edges} applies the Canny edge detector to each image \cite{ding2019sensitivity}. The corruptions of shot noise, impulse noise, glass blur, motion blur, brightness, fog, and spatter are modified from Imagenet-C \cite{hendrycks2018benchmarking}.

To construct MNIST-C we started with a broad suite of 31 image corruptions drawn from prior literature both on robustness and image processing. These corruptions range from different kinds of additive noise, blurring, digital corruptions, geometric transformations, to superimposed zigzags and squiggles. For each corruption we parameterize the severity and then choose a severity level that degrades model performance while preserving the semantic content. To choose a good set of corruptions, we first sought to understand model behavior under these corruptions. To that end we evaluated the performance of convolutional neural networks on these corruptions through extensive data augmentation experiments involving various combinations of all 31 corruptions. These experiments allowed us to choose a diverse set of corruptions which represent many different classes of model failures, and avoid picking correlated groups of corruptions on which model performance is very similar. We choose the MNIST-C corruptions with the following principles in mind:
\begin{itemize}[noitemsep]
    \item Non-triviality
    \item Semantic Invariance
    \item Realism
    \item Breadth
\end{itemize}

{\bf Non-triviality:} Each corruption should degrade the testing accuracy of various models. We tuned the severity of each corruption to a level that exposes blind spots of modern convolutional networks. As shown in a later section, we demonstrate that the error rates of CNNs increase by up to 1000\% (relative to clean MNIST error rates) when tested on MNIST-C. Furthermore, we show that our benchmark cannot be solved by naive data augmentation, nor prior methods from the adversarial defense literature. 

{\bf Semantic preservation:} As our corruptions attempt to measure failures in computer vision systems, it is critical that the perceived label of corrupted images remains invariant to a human subject. We verify this by thorough visual inspection, and we include in the appendix a random sample of images mis-classified by a simple CNN trained on standard MNIST.

{\bf Realism:} We took care to include corruptions which models could plausibly encounter in the wild, not necessarily in an adversarial setting. Our MNIST-C corruptions might occur through real-world perturbations to the camera setup (shot noise, impulse noise, motion blur, shear, scale, rotate, translate), environmental factors (brightness, stripe, fog, glass blur), or physical modification (spatter, dotted line, zigzag, Canny edges).

{\bf Breadth:} We also paid attention to the breadth of our corruptions. From an original working list of 31 corruptions we selected 15 on the criteria of covering a wide swath of possible corruptions, and also of avoiding redundancy both in visual terms and in terms of overlap in performance degradation of the models we tested. We present these 15 corruptions as our MNIST-C benchmark, though we release the source code for all 31 corruptions and welcome further work to use a different subset.

We will release both the source code for the corruptions as well as the static, pre-computed MNIST-C dataset that we used to evaluate various models, as well the algorithms used to generate the corruptions are not optimized and unsuitable for direct use in a training routine.

\section{Experimental Results}

\begin{figure*}[h]
    \centering
    \includegraphics[width=0.77\textwidth]{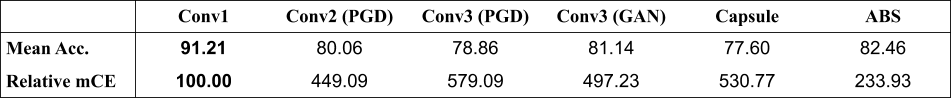}
    \caption{Mean test accuracy and relative mean Corruption Error (mCE) of various models on MNIST-C. Best performance is \textbf{bolded}.}
    \label{fig:summary}
\end{figure*}

We evaluate the MNIST-C corruptions against several models: a simple CNN (Conv1) trained on clean MNIST\footnote{Model definition taken from \href{https://github.com/pytorch/examples/blob/master/mnist/main.py}{here}.}, a different CNN (Conv2) trained against PGD adversarial noise \cite{madry2017towards}, yet another CNN (Conv3) trained against PGD/GAN adversaries \cite{wang2018direct}, a capsule network \cite{frosst2018darccc}, and a generative model, ABS \cite{schott2018robust}. The results are shown in Table~\ref{fig:main}. We find that the baseline model Conv1 achieves 91.21\% accuracy when averaged over the entire benchmark, or a 1100\% increased error rate relative to clean test accuracy (99.22\%). In the Appendix we show a random sample of test errors made by Conv1 on each corruption. We find that a majority of model errors on MNIST-C are images which are easy for a human to classify correctly. 

A popular robustness metric in recent literature has been \emph{adversarial robustness}, where model performance is evaluated against a suite of constrained optimization algorithms. It is natural to ask whether or not methods which claim improved adversarial robustness improve out-of-distribution robustness, as there are theoretical connections between $l_2$-robustness and robustness to Gaussian noise \cite{ford2019adversarial}, and prior work has observed that adversarial training improves robustness on CIFAR-10-C. To that end we investigated several previously published adversarial defenses on our benchmark. Remarkably, we find that all of the tested methods methods actually \emph{increase} the error rate on the benchmark relative to the clean model, we find that mean accuracy on MNIST-C of the three adversarially trained models is significantly lower rate of the clean model. Translating accuracy values to error rate shows the three adversarially trained models are 2.1-2.4 $\times$ as prone to error as the clean model. The two alternative architectures, the capsule network and the generative ABS model, also suffer performance degradation relative to the baseline.

A more sophisticated metric, relative mean corruption error (relative mCE), was proposed to measure performance on Imagenet-C \cite{hendrycks2018benchmarking}. Given a classifier, $f$, a baseline classifier $b$, and a single corruption $c$, the CE (corruption error) of $f$ on $c$ is computed as the ratio between $E^f_c$, the error rate of $f$ on $c$, and $E^b_c$, the error rate of $b$ on $c$. Similarly, relative CE is calculated from the ratio of the change in error of $f$ when it is evaluated on $c$ instead of the clean data, and the change in error of $b$ when it is evaluated on $c$ instead of the clean data:

$$\textrm{relative CE}^f_c = \frac{E^f_c - E^f_i}{E^b_c - E^b_i}$$

where $i$ indicates the identity corruption that returns the input image. Given these per-corruption CE and relative CE values, we then take the average across all corruptions in the dataset to compute mean CE and relative mean CE. 

Relative mCE tells a similar story to mean accuracy in figure \ref{fig:summary}. Again, across the adversarially trained CNNs and the alternative architectures we find large degradations in testing performance, as quantified by the increases in relative mCE.

The fact that these methods actually degrade OOD robustness measured through MNIST-C underscores the necessity of evaluating future methods on a broader test suite in order to quantify the robustness of a model. It is not so surprising adversarial training degrades performance on MNIST-C despite the fact it dramatically improves performance on CIFAR-10-C. Prior work has observed that the $l_\infty$-robust model on MNIST \cite{madry2017towards} achieves robustness by thresholding the input around .5, taking advantage of the fact that MNIST pixel values are concentrated near 0 and 1 \cite{schott2018robust}. Because none of our corruptions are constrained to a small $l_p$-ball, evaluating on our benchmark can easily detect undesirable solutions which overfit to the $l_p$-robustness metric.

As our benchmark is designed to measure \textit{out-of-distribution} robustness, training models directly on the distributions in the benchmark defeats its purpose. To demonstrate this, we trained and tested model Conv1 on simple mixtures of the 31 corruptions initially came up with (see Table~\ref{fig:data_aug}). We see that one can trivially recover full accuracy on any single corruption by simply finetuning\footnote{Finetuning here simply means that we pre-train on clean MNIST to convergence before switching to the corrupted training set.} on that corruption. We performed a second data augmentation experiment where we finetuned Conv1 on all but 1 one of the 31 corruptions, and then tested on the held out corruption. Remarkably we found that on average, training on all but 1 corruption only improved robustness on the remaining hold-out corruption from 90.4\% accuracy to 91.7\% accuracy, still far below the human level performance. In a flagrant violation of ``out-of-distribution'' robustness, we find that fine-tuning on all 31 corruptions together improves mean accuracy to 98.0\%.

From this experiment we draw two conclusions. First, an OOD benchmark loses its meaning when models are trained directly on the corruptions--- doing so would dramatically overestimate the robustness of a model. Second, while data augmentation can be a useful tool in improving robustness and generalization \cite{geirhos2018imagenet, cubuk2018autoaugment}, the problem of generalizing to out-of-distribution data remains highly nontrivial even when aggressive data augmentation is used. 

\section{Conclusion}

We present MNIST-C, a new robustness benchmark in computer vision. We demonstrate that our benchmark exposes new model failures that metrics in the adversarial robustness cannot detect. We note that not only does our dataset measure a much more comprehensive notion of robustness, it also reduces the difficulty of reproducibly evaluating robustness. This stands in contrast to current state of measuring adversarial robustness where reported robustness measurements are continuously refuted \cite{carlini2019evaluating,athalye2018obfuscated}.

This, in our opinion, makes MNIST-C (and the related CIFAR-10-C, and Imagenet-C) a more reliable benchmark for measuring scientific progress of robustness in computer vision. We hope that our benchmark serves as a useful tool for future work.

\Urlmuskip=0mu plus 1mu\relax
\bibliography{main}

\begin{thebibliography}{18}
\providecommand{\natexlab}[1]{#1}
\providecommand{\url}[1]{\texttt{#1}}
\expandafter\ifx\csname urlstyle\endcsname\relax
  \providecommand{\doi}[1]{doi: #1}\else
  \providecommand{\doi}{doi: \begingroup \urlstyle{rm}\Url}\fi

\bibitem[Athalye et~al.(2018)Athalye, Carlini, and
  Wagner]{athalye2018obfuscated}
Athalye, A., Carlini, N., and Wagner, D.
\newblock Obfuscated gradients give a false sense of security: Circumventing
  defenses to adversarial examples.
\newblock \emph{arXiv preprint arXiv:1802.00420}, 2018.

\bibitem[Azulay \& Weiss(2018)Azulay and Weiss]{azulay2018deep}
Azulay, A. and Weiss, Y.
\newblock Why do deep convolutional networks generalize so poorly to small
  image transformations?
\newblock \emph{arXiv preprint arXiv:1805.12177}, 2018.

\bibitem[Carlini et~al.(2019)Carlini, Athalye, Papernot, Brendel, Rauber,
  Tsipras, Goodfellow, and Madry]{carlini2019evaluating}
Carlini, N., Athalye, A., Papernot, N., Brendel, W., Rauber, J., Tsipras, D.,
  Goodfellow, I., and Madry, A.
\newblock On evaluating adversarial robustness.
\newblock \emph{arXiv preprint arXiv:1902.06705}, 2019.

\bibitem[Cubuk et~al.(2018)Cubuk, Zoph, Mane, Vasudevan, and
  Le]{cubuk2018autoaugment}
Cubuk, E.~D., Zoph, B., Mane, D., Vasudevan, V., and Le, Q.~V.
\newblock Autoaugment: Learning augmentation policies from data.
\newblock \emph{arXiv preprint arXiv:1805.09501}, 2018.

\bibitem[Ding et~al.(2019)Ding, Lui, Jin, Wang, and Huang]{ding2019sensitivity}
Ding, G.~W., Lui, K. Y.~C., Jin, X., Wang, L., and Huang, R.
\newblock On the sensitivity of adversarial robustness to input data
  distributions.
\newblock \emph{arXiv preprint arXiv:1902.08336}, 2019.

\bibitem[Dodge \& Karam(2017)Dodge and Karam]{dodge2017study}
Dodge, S. and Karam, L.
\newblock A study and comparison of human and deep learning recognition
  performance under visual distortions.
\newblock In \emph{2017 26th international conference on computer communication
  and networks (ICCCN)}, pp.\  1--7. IEEE, 2017.

\bibitem[Engstrom et~al.(2017)Engstrom, Tran, Tsipras, Schmidt, and
  Madry]{engstrom2017rotation}
Engstrom, L., Tran, B., Tsipras, D., Schmidt, L., and Madry, A.
\newblock A rotation and a translation suffice: Fooling cnns with simple
  transformations.
\newblock \emph{arXiv preprint arXiv:1712.02779}, 2017.

\bibitem[Ford et~al.(2019)Ford, Gilmer, Carlini, and
  Cubuk]{ford2019adversarial}
Ford, N., Gilmer, J., Carlini, N., and Cubuk, D.
\newblock Adversarial examples are a natural consequence of test error in
  noise.
\newblock \emph{arXiv preprint arXiv:1901.10513}, 2019.

\bibitem[Frosst et~al.(2018)Frosst, Sabour, and Hinton]{frosst2018darccc}
Frosst, N., Sabour, S., and Hinton, G.
\newblock Darccc: Detecting adversaries by reconstruction from class
  conditional capsules.
\newblock \emph{arXiv preprint arXiv:1811.06969}, 2018.

\bibitem[Geirhos et~al.(2018)Geirhos, Rubisch, Michaelis, Bethge, Wichmann, and
  Brendel]{geirhos2018imagenet}
Geirhos, R., Rubisch, P., Michaelis, C., Bethge, M., Wichmann, F.~A., and
  Brendel, W.
\newblock Imagenet-trained cnns are biased towards texture; increasing shape
  bias improves accuracy and robustness.
\newblock \emph{arXiv preprint arXiv:1811.12231}, 2018.

\bibitem[Ghifary et~al.(2015)Ghifary, Bastiaan~Kleijn, Zhang, and
  Balduzzi]{ghifary2015domain}
Ghifary, M., Bastiaan~Kleijn, W., Zhang, M., and Balduzzi, D.
\newblock Domain generalization for object recognition with multi-task
  autoencoders.
\newblock In \emph{Proceedings of the IEEE international conference on computer
  vision}, pp.\  2551--2559, 2015.

\bibitem[Goodfellow et~al.(2014)Goodfellow, Shlens, and
  Szegedy]{goodfellow2014explaining}
Goodfellow, I.~J., Shlens, J., and Szegedy, C.
\newblock Explaining and harnessing adversarial examples.
\newblock \emph{arXiv preprint arXiv:1412.6572}, 2014.

\bibitem[Hendrycks \& Dietterich(2018)Hendrycks and
  Dietterich]{hendrycks2018benchmarking}
Hendrycks, D. and Dietterich, T.~G.
\newblock Benchmarking neural network robustness to common corruptions and
  surface variations.
\newblock \emph{arXiv preprint arXiv:1807.01697}, 2018.

\bibitem[Madry et~al.(2017)Madry, Makelov, Schmidt, Tsipras, and
  Vladu]{madry2017towards}
Madry, A., Makelov, A., Schmidt, L., Tsipras, D., and Vladu, A.
\newblock Towards deep learning models resistant to adversarial attacks.
\newblock \emph{arXiv preprint arXiv:1706.06083}, 2017.

\bibitem[Pei et~al.(2017)Pei, Cao, Yang, and Jana]{pei2017towards}
Pei, K., Cao, Y., Yang, J., and Jana, S.
\newblock Towards practical verification of machine learning: The case of
  computer vision systems.
\newblock \emph{arXiv preprint arXiv:1712.01785}, 2017.

\bibitem[Schott et~al.(2018)Schott, Rauber, Brendel, and
  Bethge]{schott2018robust}
Schott, L., Rauber, J., Brendel, W., and Bethge, M.
\newblock Robust perception through analysis by synthesis.
\newblock \emph{arXiv preprint arXiv:1805.09190}, 2018.

\bibitem[Szegedy et~al.(2013)Szegedy, Zaremba, Sutskever, Bruna, Erhan,
  Goodfellow, and Fergus]{szegedy2013intriguing}
Szegedy, C., Zaremba, W., Sutskever, I., Bruna, J., Erhan, D., Goodfellow, I.,
  and Fergus, R.
\newblock Intriguing properties of neural networks.
\newblock \emph{arXiv preprint arXiv:1312.6199}, 2013.

\bibitem[Wang \& Yu(2018)Wang and Yu]{wang2018direct}
Wang, H. and Yu, C.-N.
\newblock A direct approach to robust deep learning using adversarial networks.
\newblock \emph{arXiv preprint arXiv:1905.09591}, 2018.

\end{thebibliography}
\bibliographystyle{icml2019}

\appendix
\section{Additional Corruptions}
Along with the 15 corruptions we selected for the MNIST-C benchmark suite, we also release 16 additional corruptions including. We briefly explain each of these corruptions here. 

\textbf{Speckle noise} is a random corruption which may occur during the imaging process.
\textbf{Pessimal noise} is sampled from a multivariate Gaussian with an adversarially trained covariance matrix and then tiled in a 2x2 pattern across the image. We intended this as a proxy for worst-case non-interactive corruption. The covariance matrix is trained via SGD to  maximize the training loss of model Conv1 and then frozen at test time when the corruption is applied.
\textbf{Gaussian blur} applies a Gaussian kernel to the image.
\textbf{Defocus blur} simulates a defocused camera lens.
\textbf{Zoom blur} simulates increasing focal length during image capture.
\textbf{Frost} overlays a random crop from one of six images of real frost.
\textbf{Snow} transforms and blurs Gaussian noise to simulate the appearance of snow.
\textbf{Contrast} reduces the contrast of the image.
\textbf{Saturate} increases saturation of the image.
\textbf{JPEG compression} runs the lossy JPEG compression algorithm on the image.
\textbf{Pixelate} distorts the image by resizing down and then back to the original size.
\textbf{Elastic transform} applies a random affine transformation to each square patch in the image.
\textbf{Quantize} reduces the color range of the image by rounding each pixel value to evenly spaced values.
\textbf{Line} superimposes a randomly oriented lines over the image, where the brightness of each straight segment is determined by an exponential kernel.
\textbf{Inverse} inverts the pixel values of the entire image.

Speckle noise, Gaussian blur, defocus blur, zoom blur, frost, snow, contrast, saturate, JPEG compression, pixelate, and elastic transform are taken from \cite{hendrycks2018benchmarking}.

\section{Additional Results}
\begin{figure*}[!h]
    \centering
    \includegraphics[width=0.77\textwidth]{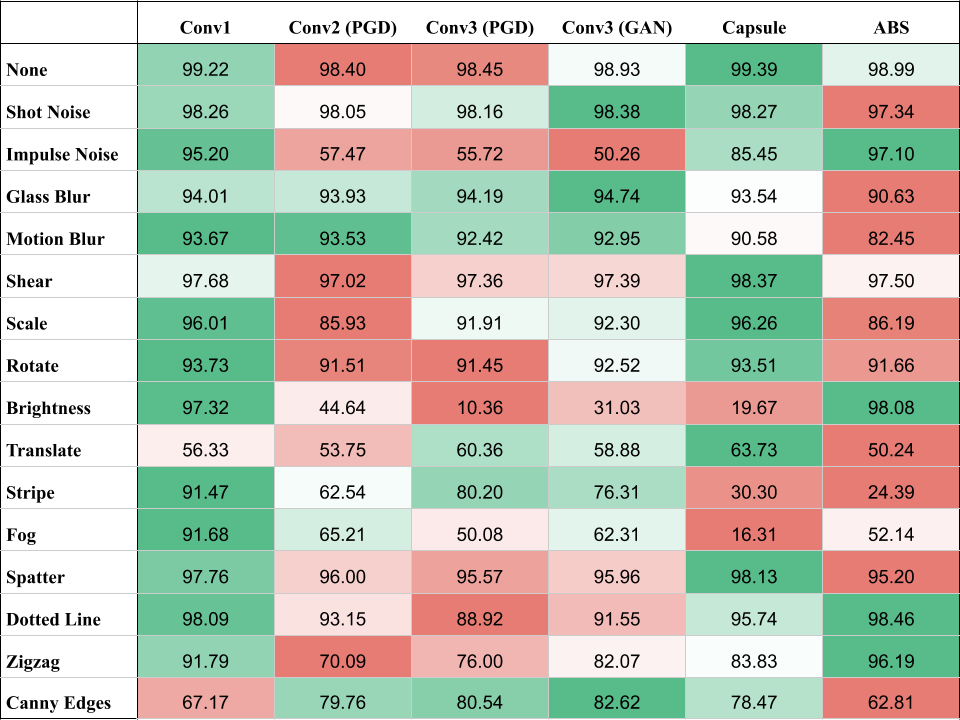}
    \caption{Test accuracy of various models on MNIST-C. Highest accuracy per row is shaded green, lowest accuracy is shaded red, and average accuracy is shaded white.}
    \label{fig:main}
\end{figure*}

We report the test accuracy of all 6 models we benchmarked on MNIST-C in figure \ref{fig:main}.

\begin{figure*}[!h]
    \centering
    \includegraphics[width=0.55\textwidth]{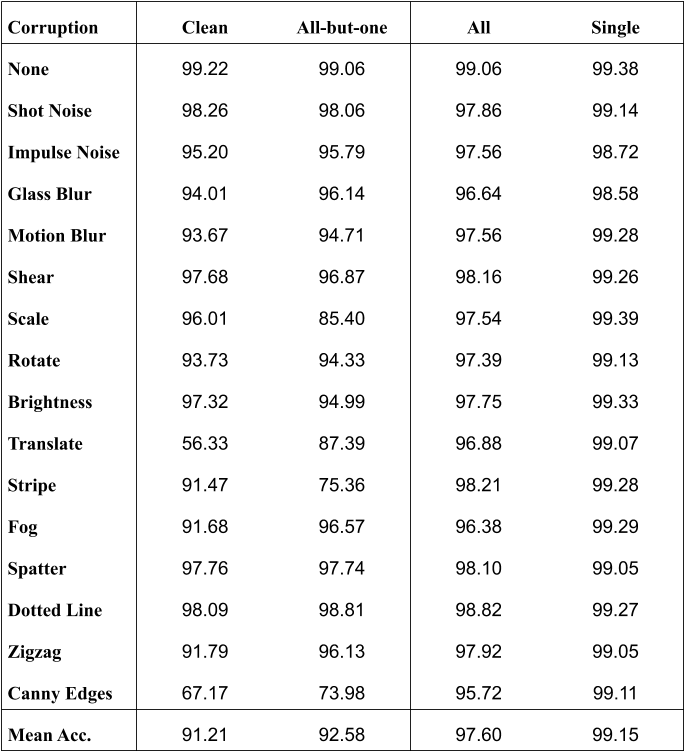}
    \caption{Test accuracy of training Conv1 with simple data augmentation methods on MNIST-C. \textbf{Clean} refers to no augmentation, \textbf{single} trains and tests on a single corruption, \textbf{all-but-one} trains on all 31 but the tested corruption, and \textbf{all} trains on all 31 corruptions.}
    \label{fig:data_aug}
\end{figure*}

In figure \ref{fig:data_aug}, we show the effects of simple data augmentation on training Conv1. The left two columns represent data augmentations that do not directly access the corruption on which the model is tested, and the right two columns represent data augmentations that do directly access the test corruption.

\section{Corruption Error Examples}
\begin{figure*}[h]
\centering
\textbf{Brightness}
\includegraphics[width=\textwidth,page=1]{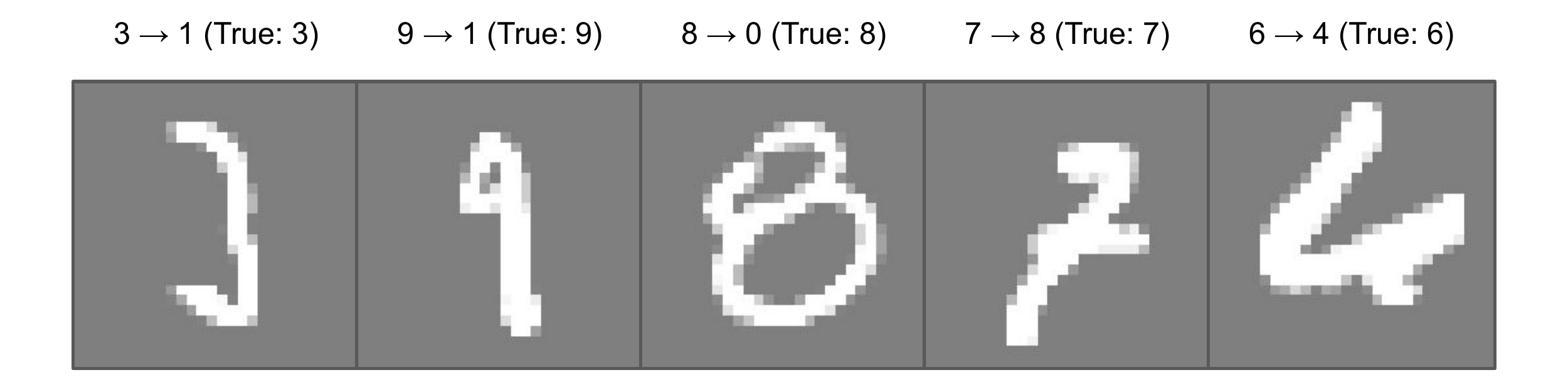}
\textbf{Canny Edges}
\includegraphics[width=\textwidth,page=2]{errors}
\textbf{Dotted Line}
\includegraphics[width=\textwidth,page=3]{errors}
\textbf{Fog}
\includegraphics[width=\textwidth,page=4]{errors}
\vspace{-5mm}
\caption{Randomly sampled test errors by Conv1 on brightness, Canny edges, dotted line, and fog.}
\label{fig:ex1}
\end{figure*}

\begin{figure*}[h]
\centering
\textbf{Impulse Noise}
\includegraphics[width=\textwidth,page=5]{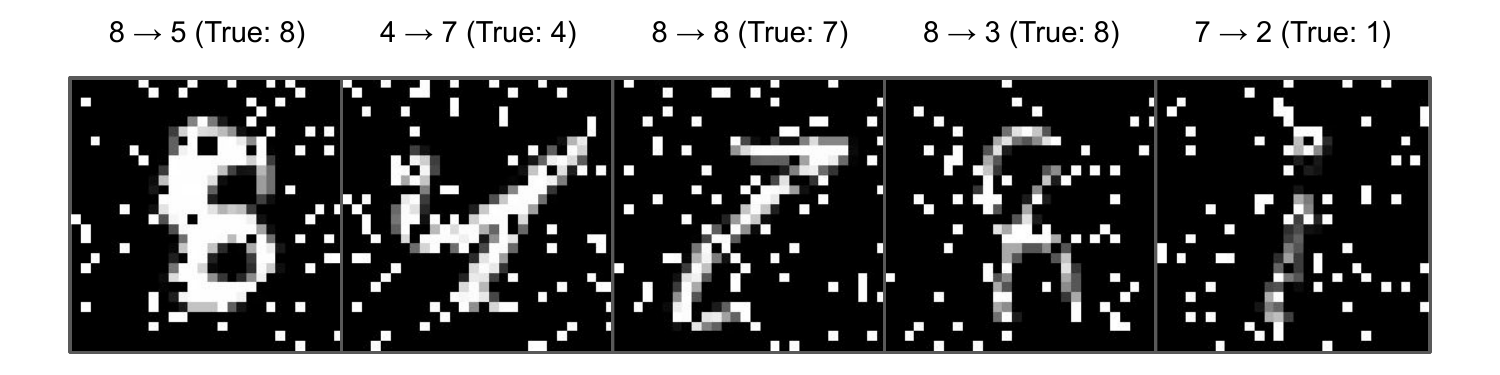}
\textbf{Stripe}
\includegraphics[width=\textwidth,page=6]{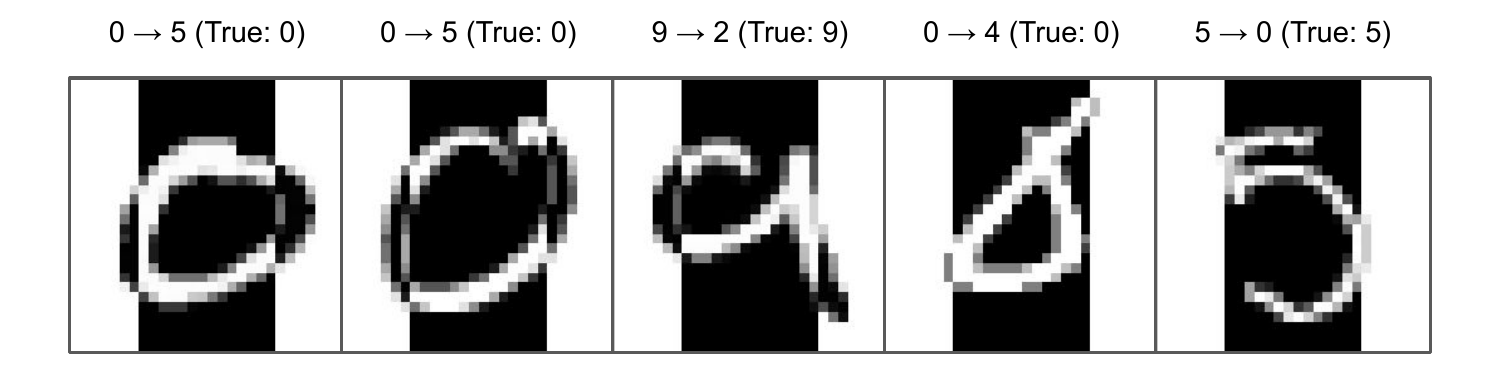}
\textbf{Motion Blur}
\includegraphics[width=\textwidth,page=7]{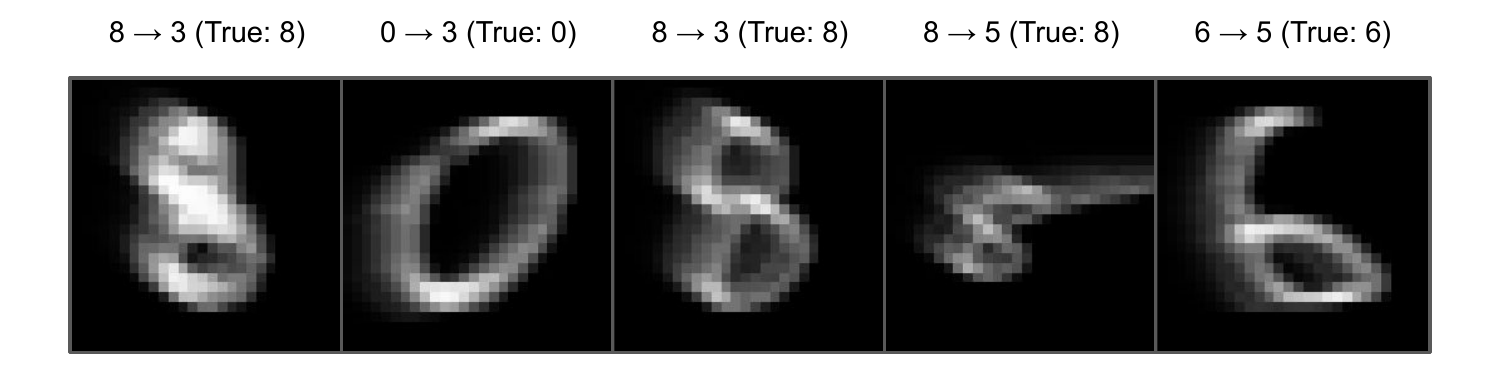}
\textbf{Glass Blur}
\includegraphics[width=\textwidth,page=8]{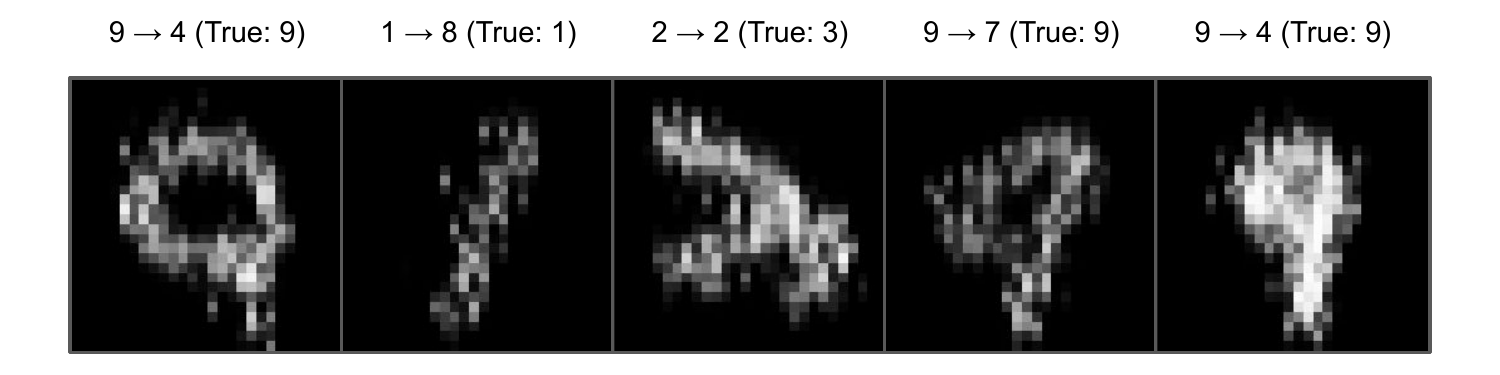}
\vspace{-5mm}
\caption{Randomly sampled test errors by Conv1 on impulse noise, stripe, motion blur, and glass blur.}
\label{fig:ex2}
\end{figure*}

\begin{figure*}[h]
\centering
\textbf{Translate}
\includegraphics[width=\textwidth,page=9]{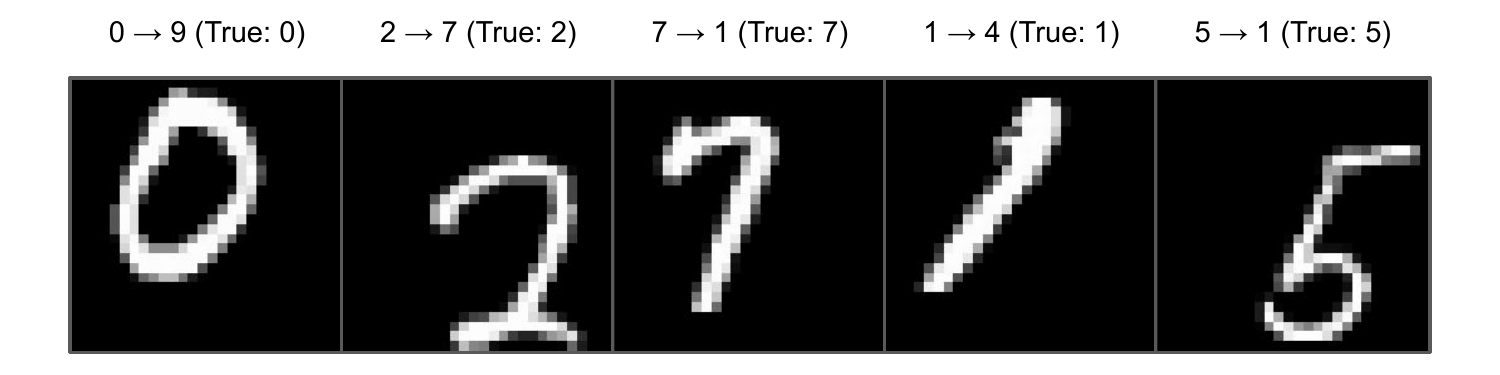}
\textbf{Rotate}
\includegraphics[width=\textwidth,page=10]{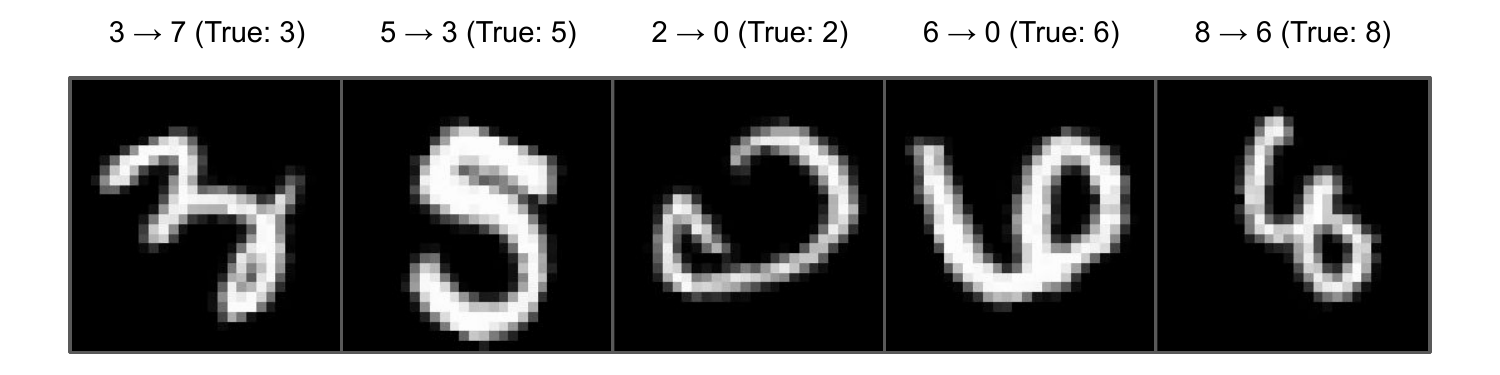}
\textbf{Scale}
\includegraphics[width=\textwidth,page=11]{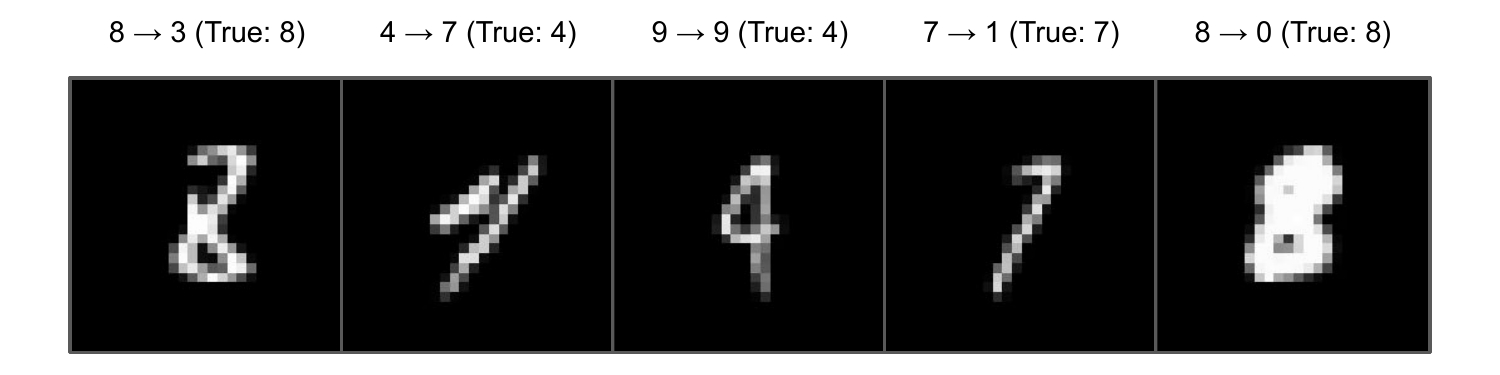}
\textbf{Shear}
\includegraphics[width=\textwidth,page=12]{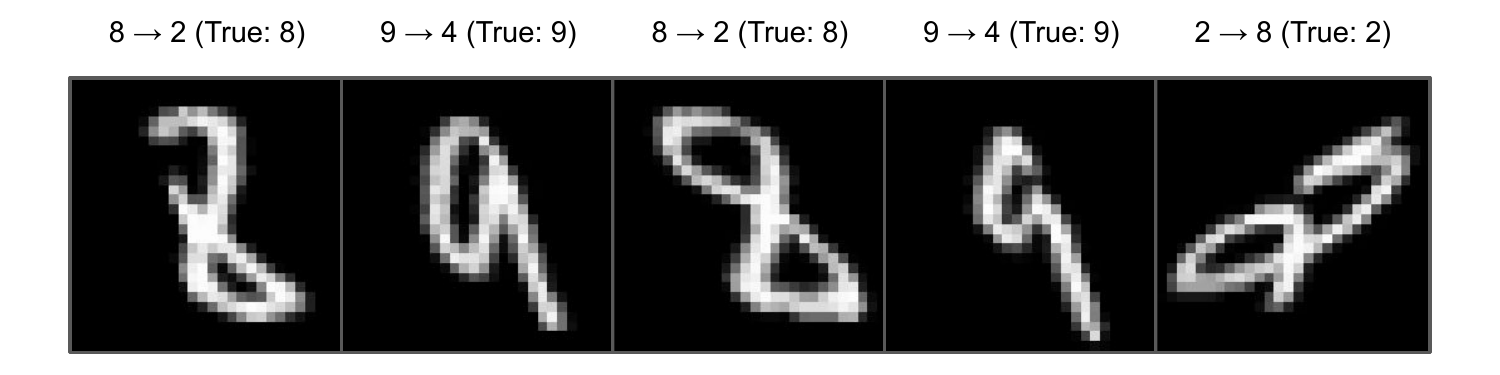}
\vspace{-5mm}
\caption{Randomly sampled test errors by Conv1 on translate, rotate, scale, and shear.}
\label{fig:ex3}
\end{figure*}

\begin{figure*}[h]
\centering
\textbf{Shot Noise}
\includegraphics[width=\textwidth,page=13]{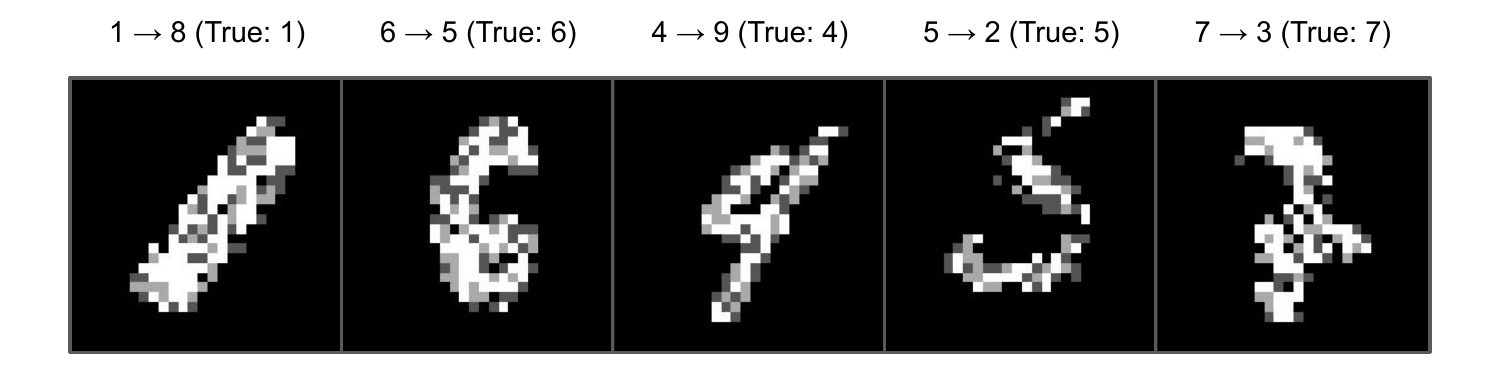}
\textbf{Spatter}
\includegraphics[width=\textwidth,page=14]{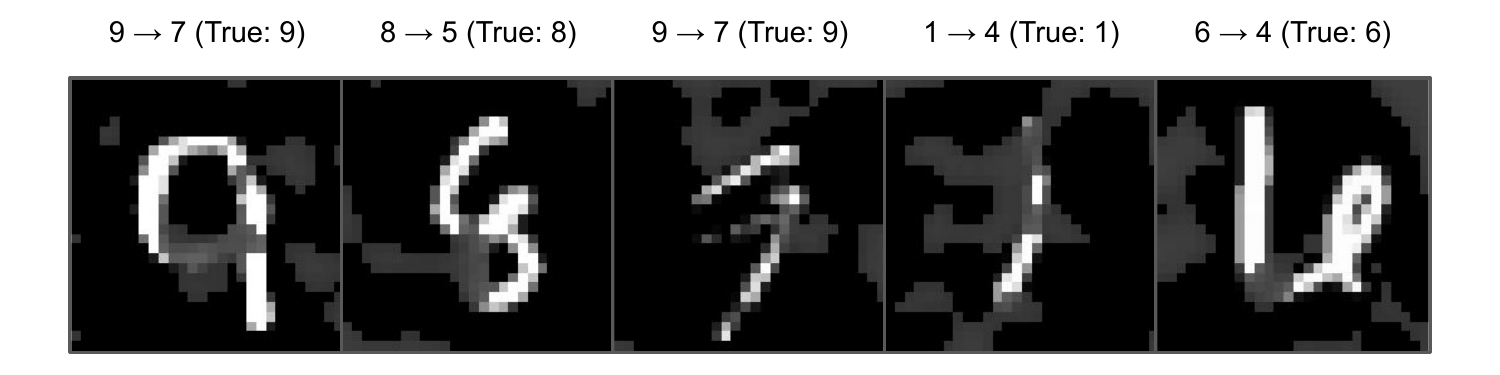}
\textbf{Zigzag}
\includegraphics[width=\textwidth,page=15]{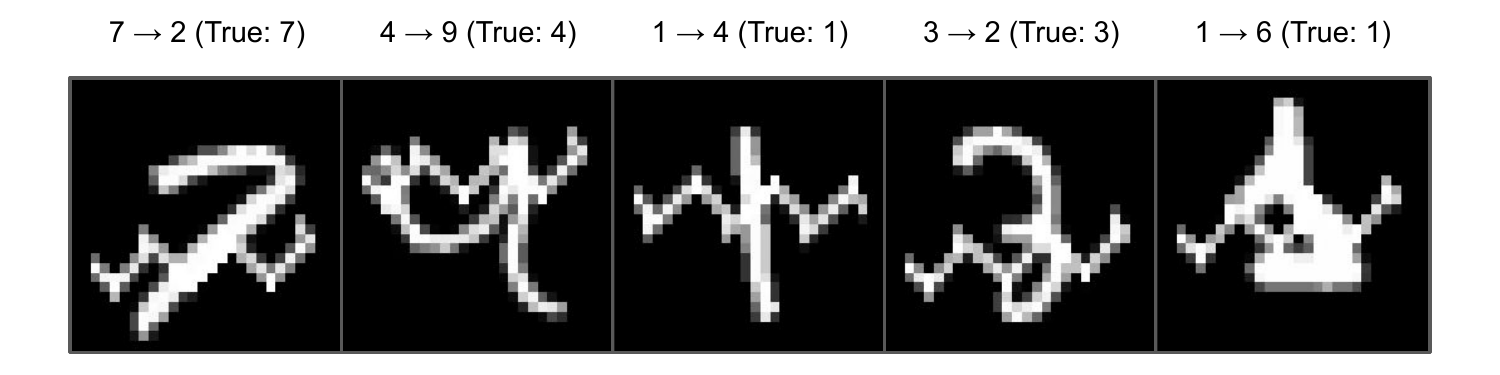}
\vspace{-5mm}
\caption{Randomly sampled test errors by Conv1 on shot noise, spatter, and zigzag.}
\label{fig:ex4}
\end{figure*}

We showcase several randomly sampled test errors made by model Conv1 on each corruption in MNIST-C. See figures \ref{fig:ex1}, \ref{fig:ex2}, \ref{fig:ex3}, \ref{fig:ex4}.

The notation $$\textrm{A} \rightarrow \textrm{B (True: C)}$$ denotes model prediction of A on the original image and B on the corrupted image, where C is true label.

\end{document}